\RequirePackage{fix-cm}

\documentclass[smallcondensed]{svjour3}

\smartqed  

\usepackage{graphicx}
\usepackage{balance}  
\usepackage{bm}
\usepackage{hyperref}
\usepackage{amsmath,stmaryrd,graphicx}
\usepackage{threeparttable}
\usepackage{multirow}
\usepackage{tikz}
\usepackage{color}
\usepackage{cite}
\usepackage{url}
\usepackage{xspace}
\usepackage{caption}
\usepackage{bold-extra}

\makeatletter

\newcommand{\thickhline}{%
    \noalign {\ifnum 0=`}\fi \hrule height 1pt
    \futurelet \reserved@a \@xhline
}
\newcommand{\fixed@sra}{$\vrule height 2\fontdimen22\textfont2 width 0pt\rightarrow$}
\newcommand{\shortarrow}[1]{%
  \mathrel{\text{\rotatebox[origin=c]{\numexpr#1*45}{\fixed@sra}}}
}
\makeatother
\definecolor{mygreen}{rgb}{0.2, 0.7, 0.2}
\definecolor{myorange}{rgb}{0.9, 0.5, 0.0}

\newcommand{\name}[1]{{\textsc{#1}}\xspace}

\newcommand{\splice}{\textsc{splice}-\textsc{junctions}\xspace}
\newcommand{\promoter}{\name{promoter}}
\newcommand{\masquerade}{\name{masquerade}}
\newcommand{\rights}{\name{rights}}
\newcommand{\unix}{\name{unix}}
\newcommand{\intrusions}{\name{intrusions}}
\newcommand{\pfam}{\name{pfam}}
\newcommand{\transfr}{\textsc{transactions}-\textsc{fr}\xspace}
\newcommand{\transmo}{\textsc{transactions}-\textsc{mo}\xspace}

\newcommand{\lstmae}{\textsc{lstm}-\textsc{ae}\xspace}
\newcommand{\lstm}{\name{lstm}}
\newcommand{\hmm}{\name{hmm}}
\newcommand{\hmms}{\textsc{hmm}s\xspace}
\newcommand{\lcs}{\name{lcs}}
\newcommand{\ism}{\name{ism}}
\newcommand{\lof}{\name{lof}}
\newcommand{\kmedoids}{\textit{k}-\textsc{medoids}\xspace}
\newcommand{\ripper}{\name{ripper}}

\newcommand{\loflcs}{\textsc{lof}-\textsc{lcs}\xspace}
\newcommand{\loflev}{\textsc{lof}-\textsc{lev}\xspace}
\newcommand{\kmedlcs}{\textit{k}-\textsc{medoids}-\textsc{lcs}\xspace}
\newcommand{\kmedlev}{\textit{k}-\textsc{medoids}-\textsc{lev}\xspace}
\newcommand{\kmed}{\textit{k}-\textsc{medoids}\xspace}
\newcommand{\knnlcs}{\textit{k}-\textsc{nn}-\textsc{lcs}\xspace}
\newcommand{\knnlev}{\textit{k}-\textsc{nn}-\textsc{lev}\xspace}
\newcommand{\tstide}{\textit{t}-\textsc{stide}\xspace}
\newcommand{\knn}{\textit{k}-\textsc{nn}\xspace}
\newcommand{\seqtoseq}{\textsc{seq}\oldstylenums{2}\textsc{seq}\xspace}

\newcommand{\uba}{\name{uba}}

\newcommand{\map}{\name{map}}
\newcommand{\ap}{\name{ap}}

\newcommand{\roc}{\name{roc}}

\begin{document}

\title{A comparative evaluation of novelty detection algorithms for discrete sequences}
\author{R\'emi Domingues \and
        Pietro Michiardi \and
        J\'er\'emie Barlet\and
        Maurizio Filippone}
\institute{R\'emi Domingues \and Pietro Michiardi \and Maurizio Filippone \\
           Department of Data Science, EURECOM, 450 Route des Chappes, Sophia Antipolis, France \\
           \{domingue, pietro.michiardi, maurizio.filippone\}@eurecom.fr \\ \\
           J\'er\'emie Barlet\\
           Amadeus, 485 Route du Pin Montard, Sophia Antipolis, France \\
           jeremie.barlet@amadeus.com}

\maketitle

\begin{abstract}
The identification of anomalies in temporal data is a core component of numerous research areas such as intrusion detection, fault prevention, genomics and fraud detection.
This article provides an experimental comparison of the novelty detection problem applied to discrete sequences.
The objective of this study is to identify which state-of-the-art methods are efficient and appropriate candidates for a given use case.
These recommendations rely on extensive novelty detection experiments based on a variety of public datasets in addition to novel industrial datasets.
We also perform thorough scalability and memory usage tests resulting in new supplementary insights of the methods' performance, key selection criterion to solve problems relying on large volumes of data and to meet the expectations of applications subject to strict response time constraints.

\end{abstract}

\section{Introduction} \label{sec:intro}
Novelty detection is an unsupervised learning problem and an active research area \cite{Aggarwal2015outlier,Hodge2004outlier}.
Given a set of training samples, novelty detection is the task of classifying new test samples as \textit{nominal} when the test data relates to the training set, or as \textit{anomalous} when they significantly differ.
Anomalous data is called novelties or anomalies and is assumed to be generated by a different generative process.
Since novelty detection can be considered a one-class classification problem, it has also been described as a semi-supervised problem\cite{Chandola2012survey} when the training set is exempt of outliers.
While most anomaly detection problems deal with numerical data\cite{emmott2016anomaly,breunig2000lof,Ramaswamy2000kNN}, novelty detection methods have been successfully applied to categorical data\cite{Hodge2004outlier}, time-series\cite{Marchi2015denoising,Zbigniew2004change,Taylor2018prophet}, discrete sequences\cite{chandola2008comparative,stide1999intrusions,Cohen1995115} and mixed data types\cite{Domingues2018DGP}.

This paper surveys the problem of detecting anomalies in temporal data, specifically in discrete sequences of events which have a temporal order.
Such a problem can be divided into two categories.
The first one is \textit{change point detection}, where datasets are long sequences in which we seek anomalous and contiguous subsequences, denoting a sudden change of behavior in the data. Use cases relating to this problem are sensor readings\cite{Zbigniew2004change} and first story detection\cite{Petrovic2010story}.
A second category considers datasets as sets of sequences, and targets the detection of anomalous sequences with respect to nominal samples. Our study focuses on the latter, which encompasses use cases such as protein identification for genomics\cite{chandola2008comparative,sun2006pst}, fraud and intrusion detection\cite{maxion2002masquerade,stide1999intrusions,chandola2008comparative} and user behavior analysis (\textsc{uba})\cite{sculley2006compression}.

While this is a matter of interest in the literature, most reviews addressing the issue focus on theoretical aspects\cite{Gupta2014survey,Chandola2012survey}, and as such do not assess and compare performance.
Chandola et al.\cite{chandola2008comparative} showcase an experimental comparison of novelty detection methods for sequential data, although this work uses a custom metric to measure the novelty detection capabilities of the algorithms and misses methods which have been recently published in the field.
Our work extends previous studies by bringing together the following contributions:
(i) comparison of the novelty detection performance for 12 algorithms, including recent developments in neural networks, on 81 datasets containing discrete sequences from a variety of research fields;
(ii) assessment of the robustness for the selected methods using datasets contaminated by outliers, with contrast to previous studies which rely on clean training data;
(iii) scalability measurements for each algorithm, reporting the training and prediction time, memory usage and novelty detection capabilities on synthetic datasets of increasing samples, sequence length and anomalies;
(iv) discussion on the interpretability of the different approaches, in order to provide insights and motivate the predictions resulting from the trained model.
To our knowledge, this study is the first to perform an evaluation of novelty detection methods for discrete sequences with so many datasets and algorithms.
This work is also the first to assess the scalability of the selected methods, which is an important selection criterion for processes subject to fast response time commitments, in addition to resource-constrained systems such as embedded systems.

The paper is organized as follows: Section \ref{sec:methods} presents the state-of-the-art of novelty detection methods, Section \ref{sec:experiments} details the real-world and synthetic datasets used for the study, in addition to the relevant metrics and parameters, Sections \ref{sec:results} and \ref{sec:conclusion} report the results and conclusions of the work.

\section{Methods} \label{sec:methods}
The current section details novelty detection methods from the literature. In order to provide recommendations relevant to real-world use cases, only methods satisfying the following constraints were selected:
(1) the method accepts \textit{discrete sequences of events} as input, where events are represented as categorical samples;
(2) the sequences fed to the method may have \textit{variable lengths}, which implies a dedicated support or a tolerance for padding;
(3) the novelty detection problem induces a distinct training and testing dataset. As such, the selected approach should be able to perform \textit{predictions on unseen data} which was not presented to the algorithm during the training phase;
(4) subject to user inputs and system changes, the set of discrete symbols in the sequences (alphabet) of the training set cannot be assumed to be complete. The algorithm should support \textit{new symbols from the test set};
(5) in order to perform an accurate evaluation of its novelty detection capabilities and to provide practical predictions on testing data, the method should provide \textit{continuous anomaly scores} rather than a binary decision. This last point allows for a ranking of the anomalies, and hence a meaningful manual validation of the anomalies, or the application of a user-defined threshold in the case of automatic intervention. The ranking of anomalies is also required by the performance metric used in the study and described in section \ref{sec:perf_datasets}.

\subsection{Hidden Markov Model}
\textbf{Hidden Markov Models} (\hmms)\cite{RabinerHMM18626} are popular graphical models used to describe temporal data and generate sequences. The approach fits a probability distribution over the space of possible sequences, and is widely used in speech recognition and protein modelling.
An \hmm is composed of $N$ states which are interconnected by state-transition probabilities, each state generating emissions according to its own emission probability distribution and the previous emission.
To generate a sequence, an initial state is first selected based on initial probabilities. A sequence of states is then sampled according to the transition matrix of the \hmm. Once the sequence of states is obtained, each state emits a symbol based on its emission distribution.
The sequence of emissions is the observed data. Based on a dataset composed of emission sequences, we can achieve the inverse process, i.e. estimate the transition matrix and the emission distributions of a \hmm from the emissions observed. Possible sequences of \textit{hidden} states leading to these emissions are thus inferred during the process.
Once we obtain a trained \hmm $\lambda = (A, B, \pi)$ with $A$ the transition matrix, $B$ describing the emission probabilities and $\pi$ the initial state probabilities, we can compute the normalized likelihood of a sequence and use it as a score to detect novelties.

\subsection{Distance-based methods}
Distance-based approaches rely on pairwise distance matrices computed by applying a distance function to each pair of input sequences. The resulting matrix is then used by clustering or nearest-neighbor algorithms to build a model of the data.
At test time, a second distance matrix is computed to perform scoring, which contains the distance between each test sample and the training data.

\subsubsection{Distance metrics}
\lcs is the \textbf{longest common subsequence}\cite{Bergroth2000Survey} shared between two sequences. A common subsequence is defined as a sequence of symbols appearing in the same order in both sequences, although they do not need to be consecutive. For example, $\lcs(\textsc{xmjyauz}, \textsc{mzjawxu}) = \textsc{mjau}$.
Since \lcs expresses a similarity between sequences, we use the negative $\lcs$ to obtain a distance.

The \textbf{Levenshtein distance}\cite{1966levenshtein}, also called the \textit{edit distance}, is a widely used metric which computes the difference between two strings or sequences of symbols. It represents the minimum number of edit operations required to transform one sequence into another, such as insertions, deletions and substitutions of individual symbols.

Both metrics are normalized by the sum of the sequence lengths (equation \ref{eq:norm_dist}), which makes them suitable for sequences of different length.

\begin{equation} \label{eq:norm_dist}
distance(x, y) = \frac{metric(x, y)}{|x| + |y|}
\end{equation}

\subsubsection{Algorithms}
The \textbf{\textit{k}-nearest neighbors} (\knn) algorithm is often used for classification and regression.
In the case of classification, \knn assigns to each test sample the label the most represented among its \textit{k} nearest neighbors from the training set.
In \cite{Ramaswamy2000kNN}, the scoring function used to detect outliers is the distance $d(x, n_k)$ or $d_k(x)$ between a point $x$ and its \textit{$k^{th}$} nearest neighbor $n_k$.
This approach was applied to sequences in \cite{chandola2008comparative} using the \lcs metric, and outperformed methods such as \hmm and \ripper.

\textbf{Local outlier factor} (\lof) \cite{breunig2000lof} also studies the neighborhood of test samples to identify anomalies.
It compares the local density of a point $x$ to the local density of its neighbors by computing the \textit{reachability distance} $rd_k(x,y)$ between $x$ and each of its \textit{k}-nearest neighbors $n_i$.
\begin{equation}
rd_k(x,n_i) = \max(d_k(n_i), d(x, n_i))
\end{equation}
The computed distances are then aggregated into a final anomaly score detailed in \cite{breunig2000lof}.
The method showed promising results when applied to intrusion detection \cite{Lazarevic2003Study}.

\textbf{\textit{k}-medoids} \cite{park2009kmedoids} is a clustering algorithm which uses data points from the training set, also called \textit{medoids}, to represent the center of a cluster.
The algorithm first randomly samples \textit{k} medoids from the input data, then cluster the remaining data points by selecting the closest medoid. The medoids of each cluster are further replaced by a data point from the same cluster which minimizes the sum of distances between the new medoid and the points in the cluster.
The method uses expectation-maximization and is very similar to \textit{k}-means, although the latter uses the arithmetic mean of a cluster as a center, called \textit{centroid}.
Since \textit{k}-means requires numerical data and is more sensitive to outliers \cite{park2009kmedoids}, it was not selected for this study.
We use the distance to the closest medoid to detect anomalies, which is the method used in \cite{Budalakoti2009airline} and \cite{budalakoti2006anomaly}.
Both papers used the \lcs metric to preprocess the data given to \kmedoids.

\subsection{Window-based techniques}
The two following methods observe subsequences of fixed length, called \textit{windows}, within a given sequence to identify abnormal patterns.
This workflow requires to preprocess the data by applying a sliding window to each sequence, shifting the window by one symbol at each iteration and resulting in a larger dataset due to overlapping subsequences.

\textbf{\tstide}\cite{stide1999intrusions}, which stands for \textit{threshold-based sequence time-delay embedding}, uses a dictionary or a tree to store subsequences of length \textit{k} observed in the training data, along with their frequency.
Once this model is built, the anomaly score of a test sequence is the number of subsequences within the sequence which do not exist in the model, divided by the number of windows in the test sequence.
For increased robustness, subsequences having a frequency lower than a given threshold are excluded from the model.
This increases the anomaly score for uncommon patterns, and allows the algorithm to handle datasets contaminated by anomalous sequences.
This scoring method is called Locality Frame Count (\textsc{lfc}) and was applied to intrusion detection\cite{stide1999intrusions} where it performed almost as well as \hmm at a reduced computational cost.

\textbf{\ripper} \cite{Cohen1995115} is a supervised classifier designed for association rule learning.
The training data given to the algorithm is divided into a set of sequences of length \textit{k}, and the corresponding labels.
For novelty detection, subsequences are generated by a sliding window, and the label is the symbol following each subsequence.
This allows \ripper to learn rules predicting upcoming events.
This method was applied to intrusion detection in \cite{lee1997learning}.
To build an anomaly score for a test sequence, the authors retrieve the predictions obtained for each subsequence, along with the confidence of the rule which triggered the prediction.
Each time a prediction does not match the upcoming event, the anomaly score is increased by $confidence * 100$. The final score is then divided by the number of subsequences for normalization.

\subsection{Pattern mining}
Sequential Pattern Mining (SPM) consists in the unsupervised discovery of interesting and relevant subsequences in sequential databases.
A recent algorithm from this field is \textbf{Interesting Sequence Miner} (\ism)\cite{Fowkes2016ISM}, a probabilistic and generative method which learns a set of patterns leading to the best compression of the database.
From a training set, \ism learns a set of interesting subsequences ranked by probability and interestingness.
To score a test sequence, we count the number of occurrences of each interesting pattern returned by \ism, and multiply the number of occurrences by the corresponding probability and interestingness.
This score is normalized by the length of the test sequence, a low score denoting an anomaly.
While alternatives to \ism exist in the literature\cite{gan2018survey}, few provide both a probabilistic framework and an open source software.

\subsection{Neural networks}
Recurrent neural networks (\textsc{rnn}s) are widely used algorithms for a variety of supervised tasks related to temporal data\cite{Lipton2015Review}.
Long Short-Term Memory (\lstm)\cite{hochreiter1997lstm}, a specific topology of \textsc{rnn}, has the ability to model long-term dependencies and thus arbitrary long sequences of events.
This network can be applied to unsupervised learning problems by using an autoencoder topology, i.e. using input and output layers of same dimensions to present the same data in input and output to the network. This allows the method to learn a compressed representation of the data.
For this purpose, the following algorithms use two multilayer \lstm networks, the first one encoding the data in a vector of fixed dimensionality (encoder), the second one decoding the target sequence from the vector (decoder).

The \textbf{Sequence to Sequence} (\seqtoseq)\cite{Sutskever2014Seq2Seq} network is a recent work designed for language translation.
The method is based on \lstm cells and uses various mechanisms such as \textit{dropout} to prevent overfitting and \textit{attention}\cite{luong2015effective} to focus on specific past events to establish correlations.
As suggested in \cite{sakurada2014anomaly,Marchi2015denoising}, the reconstruction error is used to score anomalies. The reconstruction error is the distance between the input and the reconstructed output, computed by \lcs in this study.

We also include a simpler \textbf{\lstm Autoencoder} (\lstmae) for the sake of the comparison, paired with a different scoring system.
This network is also composed of two \lstm networks, and both \seqtoseq and \lstmae perform masking to handle padding characters appended to the end of the sequences of variable length.
However, \lstmae does not benefit from the dropout and attention mechanisms.
In addition, instead of comparing the input to the reconstructed output for scoring, we now apply a distinct novelty detection algorithm to the latent representation provided by the network.
The goal of \lstmae is thus to learn a numerical fixed-length vector to represent each input sequence.
The resulting representation of the training set is given to \textbf{Isolation Forest}\cite{liu2008isolation}, an unsupervised novelty detection algorithm for numerical data recommended in \cite{emmott2016anomaly}.
At test time, the input sequence is encoded into a vector which is scored by Isolation Forest.

\section{Experimental setup} \label{sec:experiments}
\subsection{Performance tests} \label{sec:perf_datasets}
Our evaluation uses 81 datasets related to genomics, intrusion detection and user behavior analysis (\uba). The datasets are divided into 9 categories detailed in Table \ref{table:datasets}, and cover a total of 68,832 sequences.
For a given dataset, we use 70\% of the data for the training, and 30\% for the testing.

We detail thereafter the metrics used to evaluate the novelty detection capabilities of the methods.
At prediction time, each method provides us with continuous anomaly scores $s$ which allow us to rank novelties from a testing set.
We can then define a threshold $\alpha$ and classify test points as anomalies when $s > \alpha$.
The novelty detection capabilities of the algorithms can further be assessed by computing the \textit{precision} and \textit{recall} metrics on the resulting binary classification (eq. \ref{eq:precision:recall}).
These metrics require a labelled testing dataset where novelties and nominal cases are defined as \textit{positive} and \textit{negative} samples.
Data points correctly labelled as positives are called true positives (TP), examples incorrectly labelled as positives are called false positives (FP), and positive samples incorrectly labelled as negatives are referred as false negatives (FN).

\begin{equation} \label{eq:precision:recall}
precision = \frac{TP}{TP+FP}
\qquad
recall = \frac{TP}{TP+FN}
\end{equation}

By varying $\alpha$ over the range of values taken by $s$, we can compute different precision and recall measurements resulting in a precision-recall curve. The area under this curve is called average precision (\ap) and is the recommended metric to assess the performance of novelty detection methods\cite{davis2006relationship}.
An alternative metric used in the literature is the area under the receiver operating characteristic (\textsc{roc}) curve.
While the latter is widely used for classification problems, Davis et al.\cite{davis2006relationship} demonstrated that it was not appropriate when dealing with heavily imbalanced class distributions, which is inherent to novelty detection where anomalies consist in a small proportion of the labelled data.
Indeed, false positives have very little impact on the \roc, whereas \ap is strongly penalized by these, even if their proportion is not significant compared to the size of the negative class.

We thus measure the performance of the algorithms by computing the average precision (\ap) over the testing data.
To ensure stability and confidence in our results, we perform 5-fold cross-validation for each method and dataset. 
The final performance given in Table \ref{tab:map} is thus the \textit{mean average precision} (\map), i.e. the \ap averaged over the 5 iterations.
A \textit{robust} method is able to learn a consistent model from noisy data, i.e. a training set contaminated by anomalies.
We use the same proportion of outliers in the training and testing sets to showcase the robustness of the selected methods.

The corpus of data described in Table \ref{table:datasets} includes 6 widely used public collections of datasets, in addition to 3 new collections of industrial datasets from the company Amadeus.
\pfam (v31.0) describes 5 families of proteins, namely \textsc{rub} (PF00301), \textsc{tet} (PF00335), \textsc{sno} (PF01174), \textsc{nad} (PF02540) and \textsc{rvp} (PF08284).
\intrusions contains \textsc{unix} system calls for the traces \textsc{lpr-mit}, \textsc{lpr-unm}, \textsc{sendmail-cert}, \textsc{sendmail-unm}, \textsc{stide} and \textsc{xlock}.
Concerning industrial datasets, \rights details the actions performed by users in a Web application designed to manage the permissions of users and roles. The dataset shows the sessions of the 10 most active users. For each user dataset, anomalies are introduced by sampling sessions from the 9 remaining users.
\transfr and \transmo are generated from a business-oriented flight booking application and covers Web traffic coming from France and Morocco. User selection and anomaly generation were performed as described previously.


\begin{table}
\centering
\caption{Datasets benchmarked, related to genomics (\textsc{gen}), intrusion detection (\textsc{int}) or user behavior analysis (\textsc{gen}). \textit{D} is the number of datasets in each collection. The following characteristics are averaged over the collection of datasets: \textit{N} is the number of samples, \textit{A} and \textit{$p_A$} are the number and proportion of anomalies, respectively, \textit{$M_L$} is the length of the shortest sequence, \textit{$\mu_L$} is the average sequence length, \textit{$S_L$} is the entropy of the sequence lengths, \textit{$\sigma$} is the number of unique events, \textit{$S_\sigma$} is the entropy of the event distribution, \textit{$T_5$} (Top 5\%) is the proportion of events represented by the 5\% biggest events and \textit{$L_1$} (Lowest 1\%) is the proportion of the smallest events representing 1\% of the events.}
\label{table:datasets}
\resizebox{\columnwidth}{!}{
\begin{threeparttable}
\renewcommand\TPTminimum{\linewidth}
\begin{tabular}{llllllllllllll} \thickhline
\textbf{Category} & \textbf{Area} & \textbf{D} & \textbf{N} & \textbf{A} \textbf{($p_A$)} & \textbf{$M_L$} & \textbf{$\mu_L$} & \textbf{$S_L$} & \textbf{$\sigma$} & \textbf{$S_\sigma$} & \textbf{$T_5$} & \textbf{$L_1$} \\ \thickhline
\href{https://archive.ics.uci.edu/ml/datasets/Molecular+Biology+(Splice-junction+Gene+Sequences)}{\splice} & \textsc{gen} & 1 & 1710 & 55 (3.22\%) & 60 & 60 & 0.00 & 6 & 1.39 & 25.76 & 16.67 \\ 
\href{https://archive.ics.uci.edu/ml/datasets/Molecular+Biology+\%28Promoter+Gene+Sequences\%29}{\promoter} & \textsc{gen} & 1 & 59 & 6 (10.17\%) & 57 & 57 & 0.00 & 4 & 1.39 & 26.85 & 0.00 \\ 
\href{https://pfam.xfam.org/family/browse}{\pfam} & \textsc{gen} & 5 & 5166 & 165 (3.19\%) & 117 & 1034 & 0.15 & 45 & 1.17 & 83.97 & 40.00 \\ 
\href{http://www.schonlau.net/intrusion.html}{\masquerade} & \textsc{int} & 29 & 94 & 6 (6.29\%) & 100 & 100 & 0.00 & 113 & 3.40 & 49.69 & 29.55 \\ 
\href{https://www.cs.unm.edu/~immsec/systemcalls.htm}{\intrusions} & \textsc{int} & 6 & 2834 & 202 (7.14\%) & 56 & 1310 & 4.27 & 43 & 2.01 & 66.91 & 36.43 \\ 
\href{http://archive.ics.uci.edu/ml/datasets/unix+user+data}{\unix} & \uba & 9 & 1045 & 33 (3.20\%) & 1 & 31 & 3.60 & 379 & 3.31 & 77.54 & 48.86 \\ 
\rights & \uba & 10 & 677 & 22 (3.18\%) & 1 & 15 & 3.31 & 67 & 2.19 & 70.03 & 55.95 \\ 
\transfr & \uba & 10 & 215 & 7 (3.21\%) & 4 & 49 & 3.57 & 285 & 4.16 & 47.57 & 33.37 \\ 
\transmo & \uba & 10 & 386 & 12 (3.19\%) & 5 & 37 & 3.88 & 416 & 4.18 & 67.08 & 33.46 \\ \thickhline 
\end{tabular}
\end{threeparttable}
}
\end{table}

\subsection{Scalability tests} \label{sec:exp:scala}
Synthetic datasets are generated to measure the scalability of the selected methods.
Nominal data is obtained by sampling $N$ sequences of fixed length $L$ from a Markov chain. The transition matrix used by the Markov chain is randomly generated from a uniform distribution and has dimension $\sigma$, where $\sigma$ is the size of the alphabet.
Anomalies are sampled from a distinct random transition matrix of same dimension, to which we add the identity matrix. The default proportion of anomalies in the training and testing sets is 10\%.
Both transition matrices are normalized to provide correct categorical distributions.

We vary $N$, $L$ and the proportion of anomalies to generate datasets of increasing size and complexity.
We also studied the impact of $\sigma$ on the methods, and found that it had little effect on the scalability and \map.
The training time, prediction time, memory usage and novelty detection abilities of the algorithms are measured during this process.
For each configuration, we run the algorithms 3 times over distinct sampled datasets and average the metrics to increase confidence in our results.
Training and testing datasets are generated from the same two transition matrices, and have the same number of samples and outliers.

The experiments are performed on a VMWare platform running Ubuntu 14.04 LTS and powered by an Intel Xeon E5-4627 v4 CPU (10 cores at 2.6 GHz) and 256GB RAM. We use the Intel distribution of Python 3.5.2, Java 8 and R 3.3.2.
Due to the number of algorithms and the size of the datasets, we interrupt training and scoring steps lasting more than 12 hours.
Memory usage is measured by \href{https://pypi.org/project/memory_profiler/}{memory\_profiler} for algorithms written in Python and R, and by the \textsc{unix} \textit{ps} command for other languages.
We perform a garbage collection for R and Python before starting the corresponding methods.
Memory consumption is measured at intervals of $10^{-4}$ seconds, and shows the maximum usage observed during the training or scoring step.
The memory required by the plain running environment and to store the dataset is subtracted to the observed memory peak.

\subsection{Algorithms}
The implementation and configuration of the methods are detailed in Table \ref{table:algos_imp}. Parameter selection was achieved by grid-search and maximizes the \map averaged over the testing datasets detailed in Section \ref{sec:perf_datasets}.
We use \href{https://rpy2.readthedocs.io}{rpy2} to run algorithms written in R from Python, and create dedicated subprocesses for Java and C.

\begin{table}[h]
\centering
\caption{Parameters and implementations of the selected algorithms.}
\label{table:algos_imp}
\begin{threeparttable}
\resizebox{\linewidth}{!}{
\renewcommand\TPTminimum{0.5\linewidth}
\begin{tabular}{lll} \thickhline
\textbf{Algorithm} & \textbf{Language} & \textbf{Parameters}\\ \thickhline
\href{https://github.com/hmmlearn/hmmlearn}{\hmm}~\tnote{1} & Python & $components=3, iters=30, tol=10^{-2}$ \\
\href{http://mlpy.sourceforge.net/docs/3.5/lcs.html}{\lcs} & Python & n/a \\
\href{https://pypi.org/project/leven/}{Levenshtein} & Python & n/a \\
\href{http://scikit-learn.org/stable/modules/generated/sklearn.neighbors.KNeighborsClassifier.html}{\knn} & Python & $k=\max(n*0.1, 20)$ \\
\href{http://scikit-learn.org/stable/modules/generated/sklearn.neighbors.LocalOutlierFactor.html}{\lof} & Python & $k=\max(n*0.1, 50)$ \\
\href{https://github.com/letiantian/kmedoids/blob/master/kmedoids.py}{\kmed} & Python & $k=2$ \\
\href{http://www.cs.unm.edu/~immsec/software/stide_v1.2.tar.gz}{\tstide}~\tnote{2} & C & $k=6, t=10^{-5}$ \\
\href{https://cran.r-project.org/web/packages/RWeka/index.html}{\ripper}~\tnote{2} & R & $K=9, F=2, N=1, O=2$ \\
\href{https://github.com/mast-group/sequence-mining}{\ism} & Java & $iters=100, s=10^{5}$ \\
\href{https://www.tensorflow.org/tutorials/seq2seq}{\seqtoseq}~\tnote{3} & Python & $iters=100, batch=128, hidden=40, enc\_dropout=0.5, dec\_dropout=0.$ \\
\lstmae~\tnote{3} & Python & $batch=128, iters=50, hidden=40, \delta=10^{-4}$ \\ \thickhline
\end{tabular}
}
\begin{tablenotes}
\scriptsize
\item [1] New symbols are not supported natively by the method.
\item [2] Sequences were split into sliding windows of fixed length.
\item [3] Padding symbols were added to the datasets to provide batches of sequences having the same length.
\end{tablenotes}
\end{threeparttable}
\end{table}

\section{Results} \label{sec:results}
\subsection{Novelty detection capabilities} \label{sec:perf}
The mean average precision (\map) resulting from the experiment detailed in Section \ref{sec:perf_datasets} is reported in Table \ref{tab:map} for each algorithm and dataset.
When no significant difference can be observed between a given \map and the best result achieved on the dataset, we highlight the corresponding \map in bold. The null hypothesis is rejected based on a pairwise Friedman test\cite{GARCIA20102044} with a significance level of $0.05$.

While we believe that no method outperforms all others, and that each problem may require a distinct method, we attempt to give a broad overview of how methods compare to one another.
For this purpose, we extract the rank of each algorithm on each collection of datasets from Table \ref{tab:map} and aggregate them to produce an overall ranking reported in the last column.
The aggregation is performed using the Cross-Entropy Monte Carlo algorithm \cite{pihur2009rankaggreg} and rely on the Spearman distance.

\begin{table}
\centering
\caption{Mean area under the precision-recall curve (\map) averaged per group of datasets over 5 cross-validation iterations.
Results in bold indicate that we cannot reject the null hypothesis of the given \map to be identical to the best \map achieved for the dataset. Column \textit{Rank} reports the aggregated rank for each method based on the Spearman footrule distance.}
\label{tab:map}
\resizebox{\columnwidth}{!}{
\begin{threeparttable}
\renewcommand\TPTminimum{\linewidth}
\setlength\tabcolsep{4.5pt}
\begin{tabular}{llllllllllll} \thickhline
    & \textsc{splice} & \textsc{promot.} & \pfam & \textsc{masque.} & \textsc{intrus.} & \unix & \rights & \textsc{trans}-\textsc{fr} & \textsc{trans}-\textsc{mo} & \textbf{Mean} & \textbf{Rank}\\ \thickhline
\hmm & 0.027 & 0.336 & 0.387 & \textbf{0.166} & \textbf{0.580} & \textbf{0.302} & \textbf{0.246} & \textbf{0.260} & \textbf{0.164} & \textbf{0.274} & \textbf{1} \\
\knnlcs & 0.032 & \textbf{0.437} & \textbf{0.516} & 0.132 & \textbf{0.425} & \textbf{0.207} & \textbf{0.270} & \textbf{0.179} & \textbf{0.097} & \textbf{0.255} & \textbf{3} \\
\knnlev & 0.033 & \textbf{0.412} & \textbf{0.516} & 0.129 & \textbf{0.405} & \textbf{0.120} & \textbf{0.188} & \textbf{0.185} & 0.083 & \textbf{0.230} & \textbf{5} \\
\loflcs & \textbf{0.042} & 0.150 & 0.029 & \textbf{0.167} & 0.141 & 0.073 & 0.042 & 0.091 & 0.041 & \textbf{0.086} & \textbf{12} \\
\loflev & 0.031 & 0.226 & \textbf{0.517} & \textbf{0.156} & 0.181 & \textbf{0.132} & \textbf{0.191} & \textbf{0.192} & \textbf{0.099} & \textbf{0.192} & \textbf{4} \\
\kmedlcs & 0.027 & \textbf{0.581} & \textbf{0.510} & 0.134 & 0.318 & \textbf{0.155} & \textbf{0.218} & \textbf{0.184} & 0.092 & \textbf{0.247} & \textbf{6} \\
\kmedlev & \textbf{0.040} & \textbf{0.692} & \textbf{0.513} & \textbf{0.148} & 0.222 & 0.086 & 0.146 & \textbf{0.189} & 0.078 & \textbf{0.235} & \textbf{7} \\
\tstide & \textbf{0.048} & \textbf{0.806} & \textbf{0.506} & 0.122 & \textbf{0.469} & 0.081 & 0.130 & 0.136 & \textbf{0.112} & \textbf{0.268} & \textbf{9} \\
\ripper & 0.028 & \textbf{0.431} & 0.034 & \textbf{0.176} & \textbf{0.359} & 0.053 & 0.077 & 0.105 & 0.079 & \textbf{0.149} & \textbf{10} \\
\ism & 0.027 & 0.205 & 0.116 & 0.140 & \textbf{0.559} & \textbf{0.220} & \textbf{0.217} & \textbf{0.211} & \textbf{0.111} & \textbf{0.201} & \textbf{2} \\
\seqtoseq & \textbf{0.072} & 0.341 & 0.035 & \textbf{0.178} & 0.113 & 0.076 & 0.083 & 0.092 & 0.063 & \textbf{0.117} & \textbf{11} \\
\lstmae & 0.034 & \textbf{0.494} & \textbf{0.591} & \textbf{0.178} & 0.174 & 0.074 & 0.100 & \textbf{0.173} & 0.075 & \textbf{0.210} & \textbf{8} \\ \thickhline
\end{tabular}
\end{threeparttable}
}
\end{table}

\begin{figure}[!htb]
\centering
\includegraphics[width=0.8\linewidth]{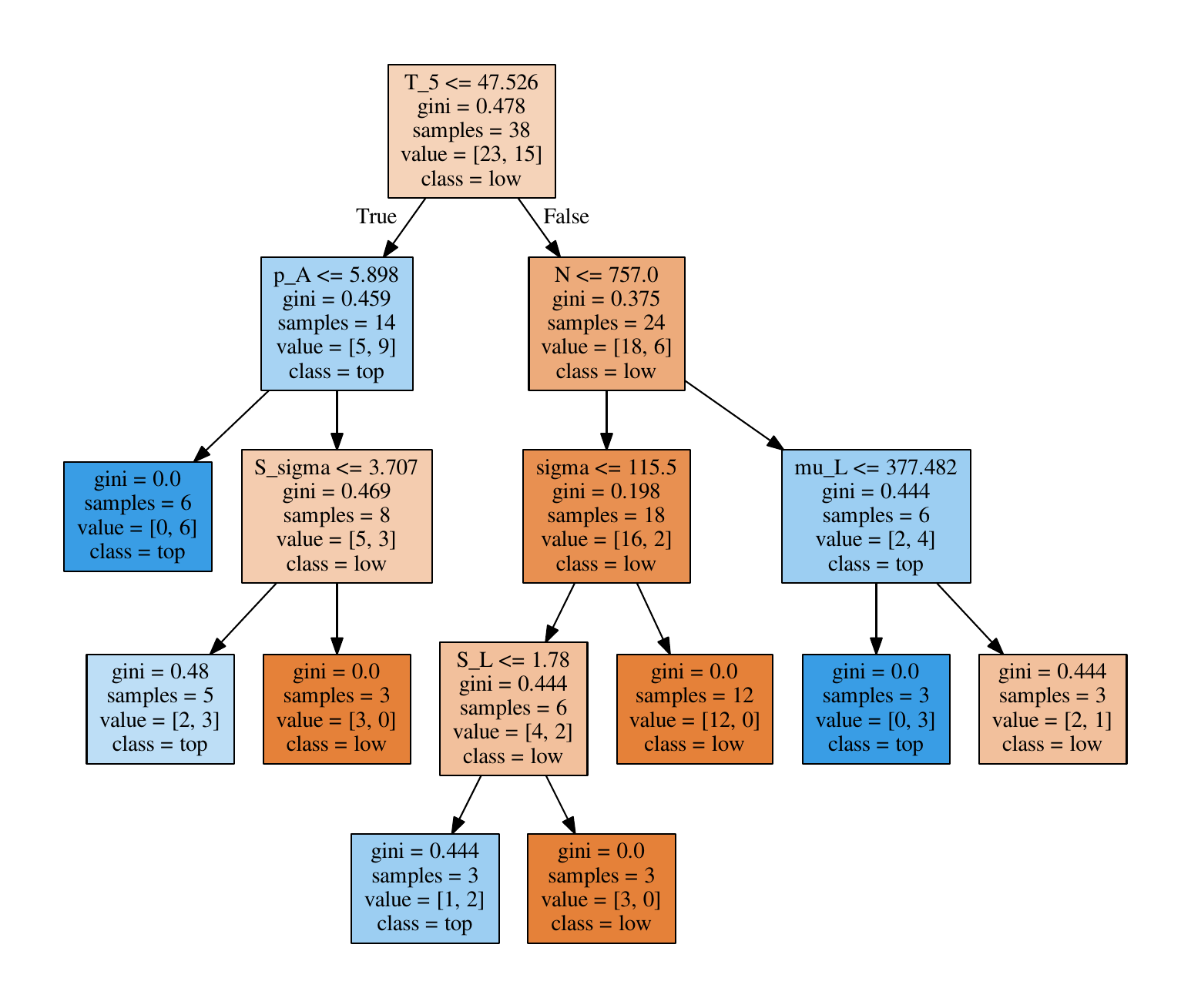}
\caption{Decision tree showing the position of \kmedlev in the overall ranking based on features extracted from the datasets. Ranks have been aggregated into the \textit{top} and \textit{low} classes which encompass the best (1 to 3) and worst (10 to 12) 25\% ranks, respectively.}
\label{fig:kMed-tree}
\end{figure}

In order to infer the behavior of each method based on the datasets characteristics, we learn an interpretable meta-model using the features introduced in Table \ref{table:datasets}.
While the metrics given in Table \ref{table:datasets} are computed over entire datasets, then averaged over the corresponding collection, this experiment focuses on the training data and retains features for each of the 81 datasets.
We use these features as input data, and fit one decision tree per algorithm in order to predict how a given method performs.
The resulting models are binary classifiers where the target class is whether the average rank of the algorithm is among the top 25\% performers (ranks 1 to 3), or if it reaches the lowest 25\% (ranks 9 to 12).
Figure \ref{fig:kMed-tree} shows the trained meta-model of \kmedlev as an example.
These trees expose the strengths and weaknesses of the methods studied, and highlight the most important factors impacting the methods' performances.

\begin{figure}[!htb]
\centering
\includegraphics[width=\linewidth]{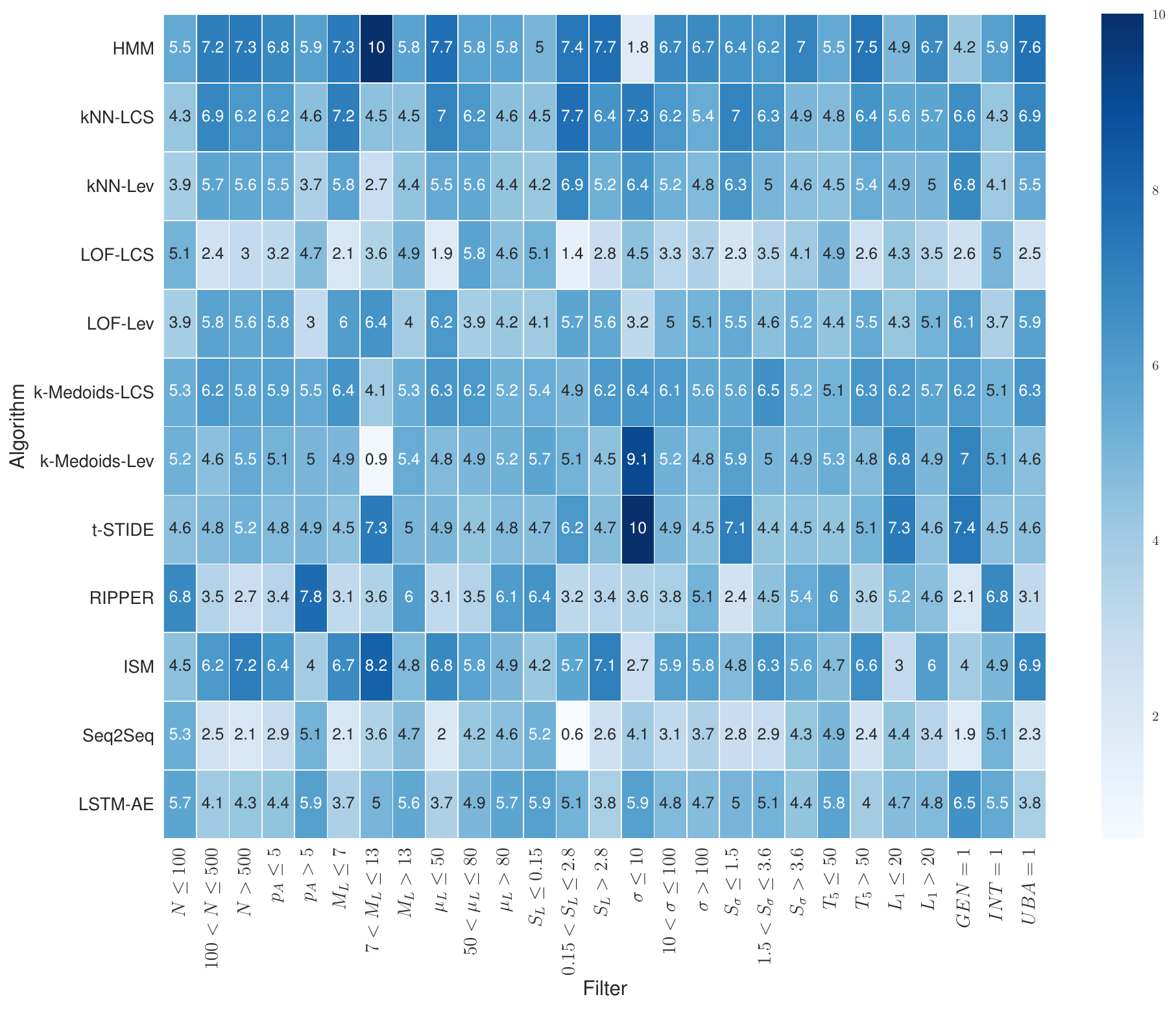}
\caption{Novelty detection capabilities of the algorithms based on the datasets characteristics. The scores range from 0 to 10 and are based on the rank of the method averaged over the subset of datasets matching the corresponding filter applied to the 81 datasets. A score of 10 corresponds to an average rank of 1, while a score of 0 indicates that the method consistently ended in the last position. $N$ is the number of samples; $p_A$ is the proportion of anomalies; $M_L$, $\mu_L$ and $S_L$ are the minimum, average and entropy computed over the sequence length; $\sigma$ and $S_\sigma$ are the alphabet size and the corresponding entropy of its distribution, the entropy increasing with the number of events and the distribution uniformity; $T_5$ is the proportion of events represented by the 5\% biggest events, a high value denotes important inequalities in the distribution; $L_1$ is the proportion of the smallest events representing 1\% of the data, a high value indicates numerous events with rare occurrences; the genomics (\textsc{gen}), intrusion detection (\textsc{int}) and \textsc{uba} columns target datasets related to the corresponding field of study.}
\label{fig:heatmap}
\end{figure}

In order to provide a concise visual overview of this analysis, we report in Figure \ref{fig:heatmap} the performance of each method based on the datasets' characteristics.
For this purpose, we extract the rules of the nodes for which $depth < 4$ in all meta-models, then aggregate these rules per feature to identify values corresponding to the most important splits.
The resulting filters are reported in the horizontal axis of the heatmap.

Our experiments show that no algorithm consistently reaches better results than the competing methods, but that \hmm, \knn and \ism are promising novelty detection methods.
While previous comparisons\cite{stide1999intrusions,chandola2008comparative,Budalakoti2009airline} use clean datasets exempt of anomalies, our study shows a good robustness for the selected methods, even for datasets with a high proportion of outliers, namely \promoter, \masquerade and \intrusions.

Concerning the applications studied, \knn, \kmedoids, \tstide and \lstmae show good performance on datasets related to genomics, which are \splice, \promoter and \pfam.
\tstide apart, these methods have successfully addressed numerous supervised numerical problems, and could thus reach good performance when applied to sequence-based supervised use cases.
The best methods for intrusion detection are \hmm and \ripper, while \tstide shows reduced performance compared to \cite{stide1999intrusions}, likely caused by the introduction of anomalies in the training sets.
Our observations for genomics and intrusion detection corroborate the conclusions presented for \tstide and \ripper in \cite{chandola2008comparative}.
However, our study shows much better performance for \hmm, the previous study using a custom likelihood for \hmm based on an aggregated sequence of binary scores.
With regard to user behavior analysis, \hmm, \knn, \kmedlcs and \ism show the best ability to differentiate users.
While the performance of \tstide on \uba is not sufficient to recommend the method, we believe that increasing the threshold of \tstide would lead to increased performance.
Indeed, user actions are often based on well-defined application flows, and most of the possible subsequences are likely to exist in the training sets.
The amount of supplementary information which can be provided by the models about the user behaviors will determine the most suitable methods for this field (Section \ref{sec:interpretability}).

Figure \ref{fig:heatmap} shows that the performance of \hmm improves significantly with the number of available samples. Both \hmm and \ism achieve good performance, even when a high discrepancy is observed among the sequence lengths.
\hmm, \ism and \ripper are able to handle efficiently a large alphabet of symbols.
\ripper also shows good performance for datasets containing a high proportion of outliers, while nearest neighbor methods are strongly impacted by this characteristic.
Distance metrics are known to suffer from the curse of dimensionality inherent to a high number of features. Similarly, Figure \ref{fig:heatmap} shows a decrease of performance for \knn, \kmedoids and \lof when $\sigma$ increases, these methods relying on the \lcs and Levenshtein metrics for distance computations.
While \lcs is a metric widely used in the literature\cite{chandola2008comparative, Budalakoti2009airline, budalakoti2006anomaly}, our experiments show that it does not perform better than the Levenshtein distance.
If both \lcs and the Levenshtein distance metrics provide satisfactory results for novelty detection when paired with \knn or \kmedoids, the combination of \lof and \lcs produces the lowest accuracy of our evaluation.
Nonetheless, the efficiency of \loflev prevents us from discarding this method, even though \knnlev achieves a similar accuracy to \loflev with a simpler scoring function.
For the sake of the experiment, we evaluated the scoring function proposed for \tstide in \cite{hofmeyr1998intrusion}. For each subsequence of fixed length in a test sequence, the authors compute the hamming distance between the test window and all training windows, and return the shortest distance. This method was much slower than a binary decision based on the presence of the test window in the training set, and did not strongly improve the results.
Neural networks do not stand out in this test. The reconstruction error showed good results for detecting numerical anomalies in previous studies\cite{sakurada2014anomaly,Marchi2015denoising}, but the approach may not be appropriate for event sequences.
The reconstructed sequences provided by \seqtoseq are often longer than the input data, and the network loops regularly for a while over a given event.
Figure \ref{fig:heatmap} show that \lstm networks perform better with long sequences and a moderate alphabet size.
We repeated our experiments using the Python library \href{https://docs.python.org/3.5/library/difflib.html#difflib.SequenceMatcher.ratio}{difflib} as an alternative to \lcs for \seqtoseq, but it did not improve the performance of the network.
\lstmae shows an acceptable novelty detection accuracy, which could be further improved with dropout and attention.
Thanks to their moderate depth, these two networks do not require very large datasets to tune their parameters. For example, \lstmae achieves a good \map even for small datasets such as \promoter and \masquerade.
Despite the use of masks to address padding, these methods have difficulty with datasets showing an important disparity in sequence length, such as \intrusions and the four collections of \uba datasets.

\subsection{Robustness}

\begin{figure}[!htb]
\centering
\includegraphics[width=0.9\linewidth]{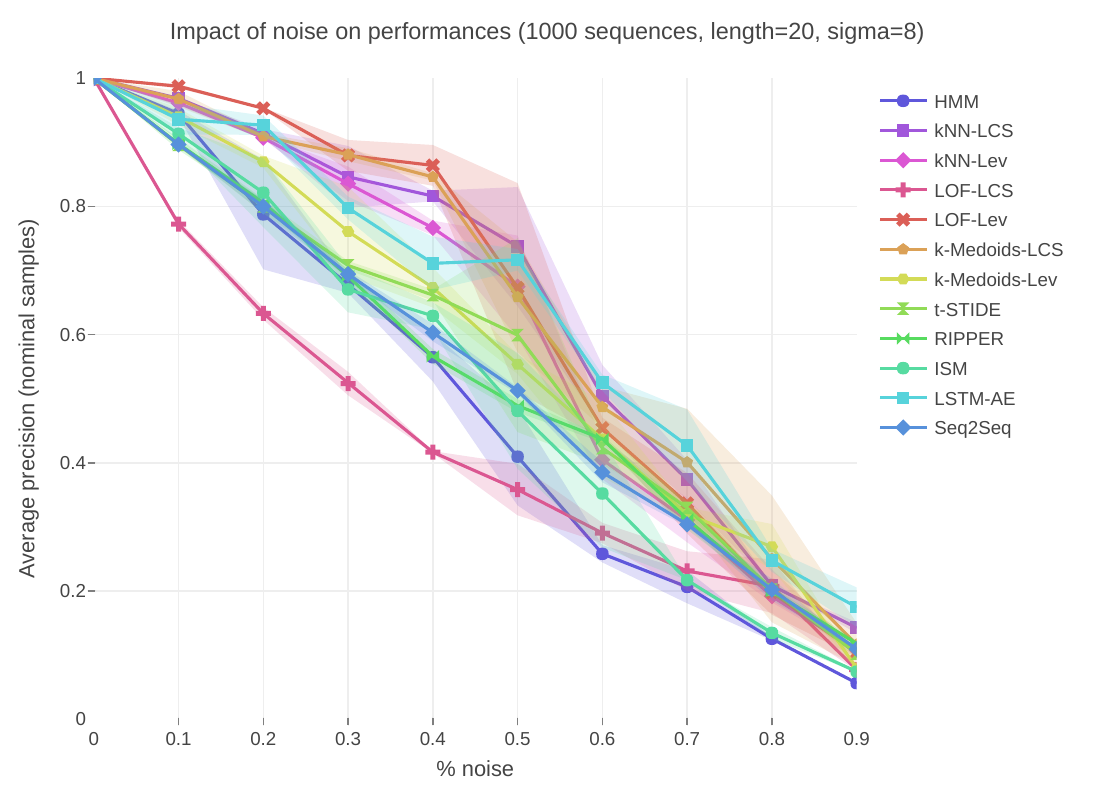}
\caption{Robustness for increasing noise density}
\label{fig:robust_outliers}
\end{figure}

\begin{figure}[!htb]
\centering
\includegraphics[width=0.9\linewidth]{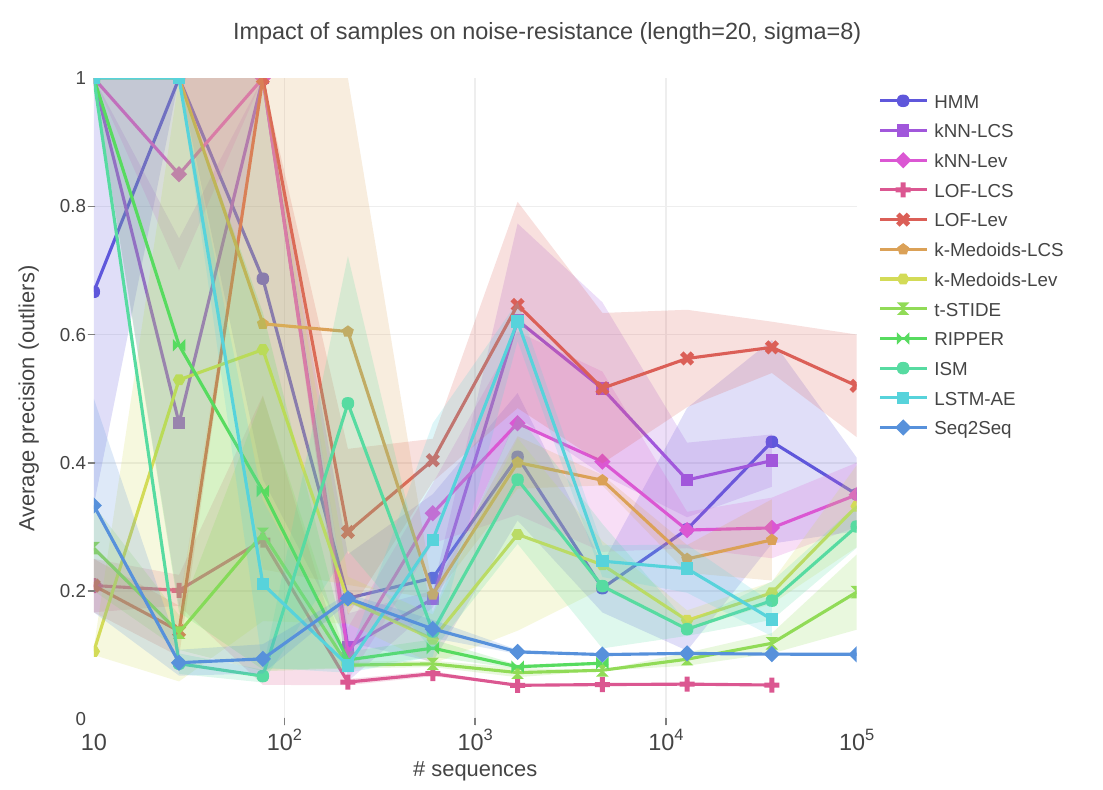}
\caption{Robustness for increasing number of samples}
\label{fig:robust_samples}
\end{figure}

\begin{figure}[!htb]
\centering
\includegraphics[width=0.9\linewidth]{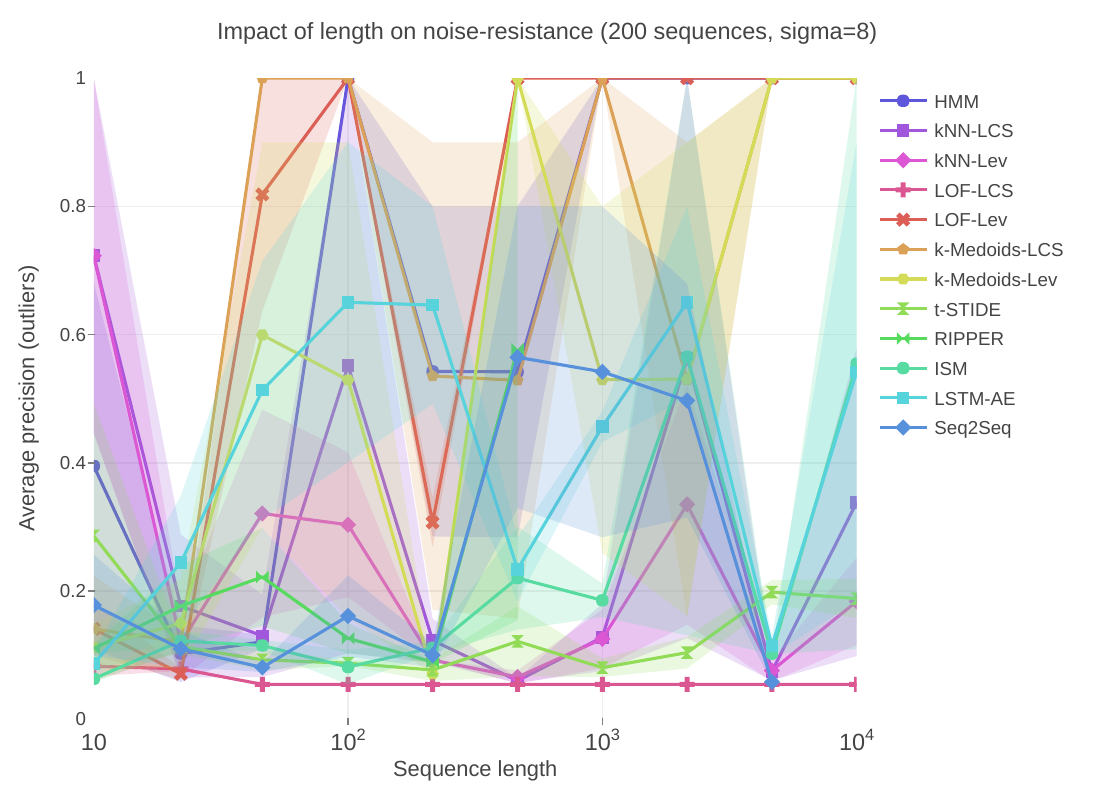}
\caption{Robustness for increasing sequence length}
\label{fig:robust_seq_len}
\end{figure}

Figures \ref{fig:robust_outliers} to \ref{fig:robust_seq_len} report the mean area under the precision recall curve (\map) for datasets of increasing proportion of outliers, number of samples and sequence length, respectively. The positive class represents the nominal samples in Figure \ref{fig:robust_outliers}, and the anomalies in Figure \ref{fig:robust_samples} and \ref{fig:robust_seq_len} (as in Section \ref{sec:perf}).

Figure \ref{fig:robust_outliers} demonstrates a more complex test case than just identifying uniform background noise against a well-defined distribution.
In this test, anomalies are sampled according to their own probability distribution, which will affect the models learnt when a sufficient proportion of anomalies is reached.
The test highlights thus how algorithms deal with complex data based on multiple distributions.
We observe that most algorithms focus on the major distribution as long as the proportion of corresponding samples remains higher than 60\%.
\hmm uses 3 components and may thus learn the second distribution much earlier in the test.
On the opposite, most of the distance-based methods discard the smallest distribution even if this one represents up to 40\% of the data.
\loflcs shows poor performance from the very beginning, which prevents us from concluding on the behavior of this method.

Figure \ref{fig:robust_samples} shows that 200 samples are a good basis to reach stable novelty detection results.
While we expected the performance of deep learning methods to improve with the number of samples, these networks did not significantly increase their detection with the size of the dataset.
The best results on large datasets were achieved by distance-based methods, most of which rely on nearest-neighbor approaches particularly efficient when a high number of samples is available.
Good performance were also achieved by \hmm, presumably due to a generation method for nominal samples and outliers based on Markov chains, which matches the internal representation of \hmm.

Despite the increasing volume of data over the scalability test reported in Figure \ref{fig:robust_seq_len}, important variations can be observed for the results, e.g. for a length of 200 events.
Such variations are probably related to the limited number of samples used in the generated datasets, which is a computational requirement to perform experiments with long sequences for several algorithms.
\kmedoids achieve better performance than other distance-based methods, which suggests a better approach for small datasets.
\hmm achieves once again good results, while \lstm networks show improved novelty detection capabilities for datasets containing sequences longer than 100 events.
The performance of \ism also increases with the volume of data, although the method require bigger datasets to reach comparable results.

In summary, our experiments show that robust models require at least 200 training samples to provide satisfactory results.
\loflcs and \tstide do not provide satisfactory performance, even though fine-tuning \tstide by increasing the frequency threshold could lead to better results.

\subsection{Runtime performance}
\begin{figure}[!htb]
\centering
\includegraphics[width=0.9\linewidth]{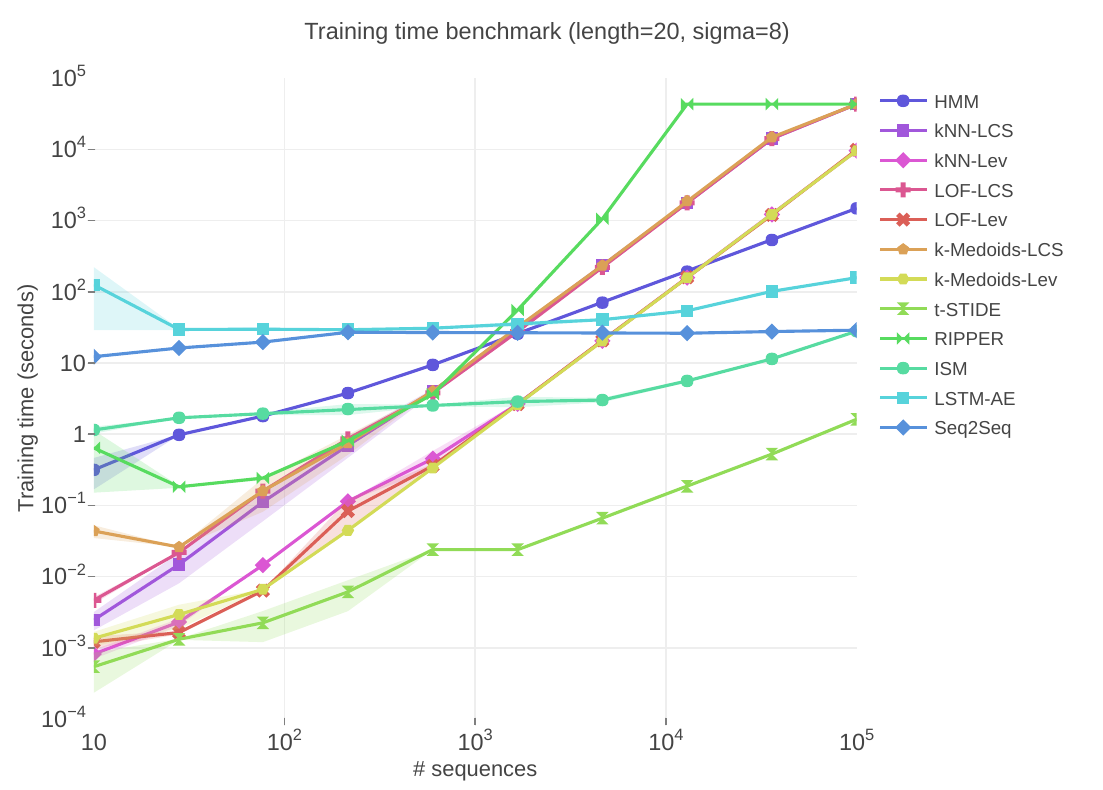}
\caption{Training time for increasing number of samples}
\label{fig:fit_samples}
\end{figure}

\begin{figure}[!htb]
\centering
\includegraphics[width=0.9\linewidth]{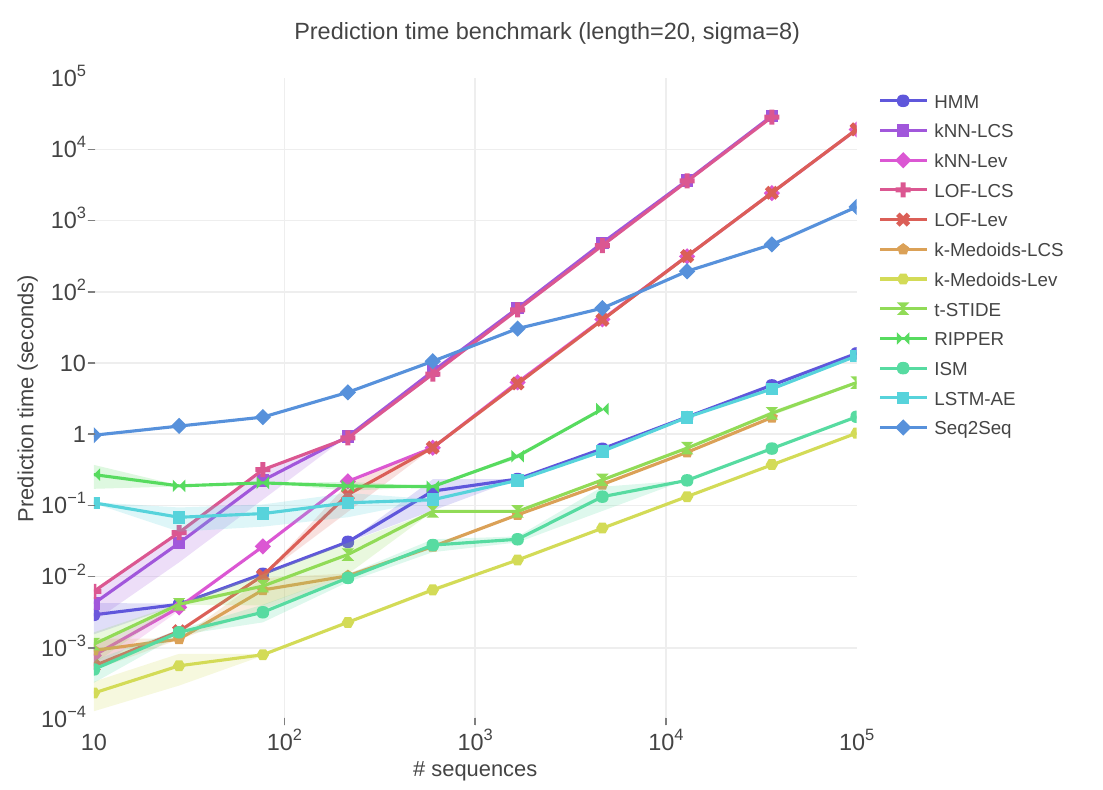}
\caption{Prediction time for increasing number of test samples}
\label{fig:pred_samples}
\end{figure}

\begin{figure}[!htb]
\centering
\includegraphics[width=0.9\linewidth]{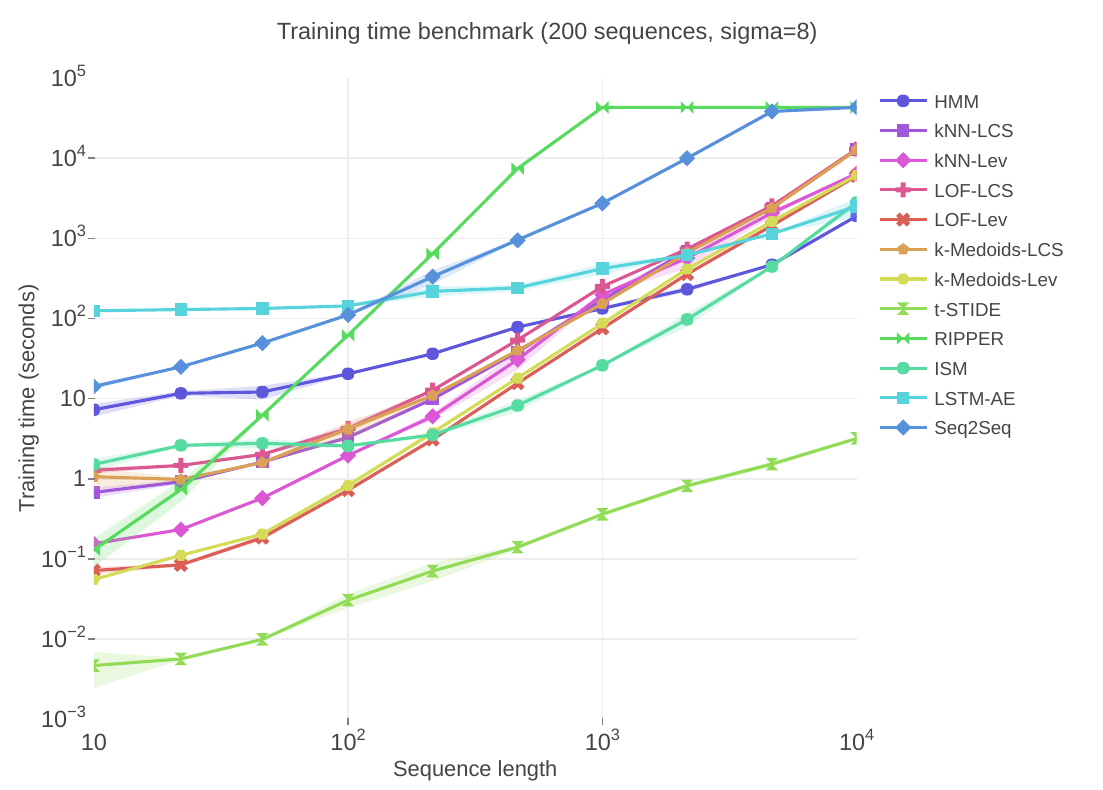}
\caption{Training time for increasing sequence length}
\label{fig:fit_seq_len}
\end{figure}

\begin{figure}[!htb]
\centering
\includegraphics[width=0.9\linewidth]{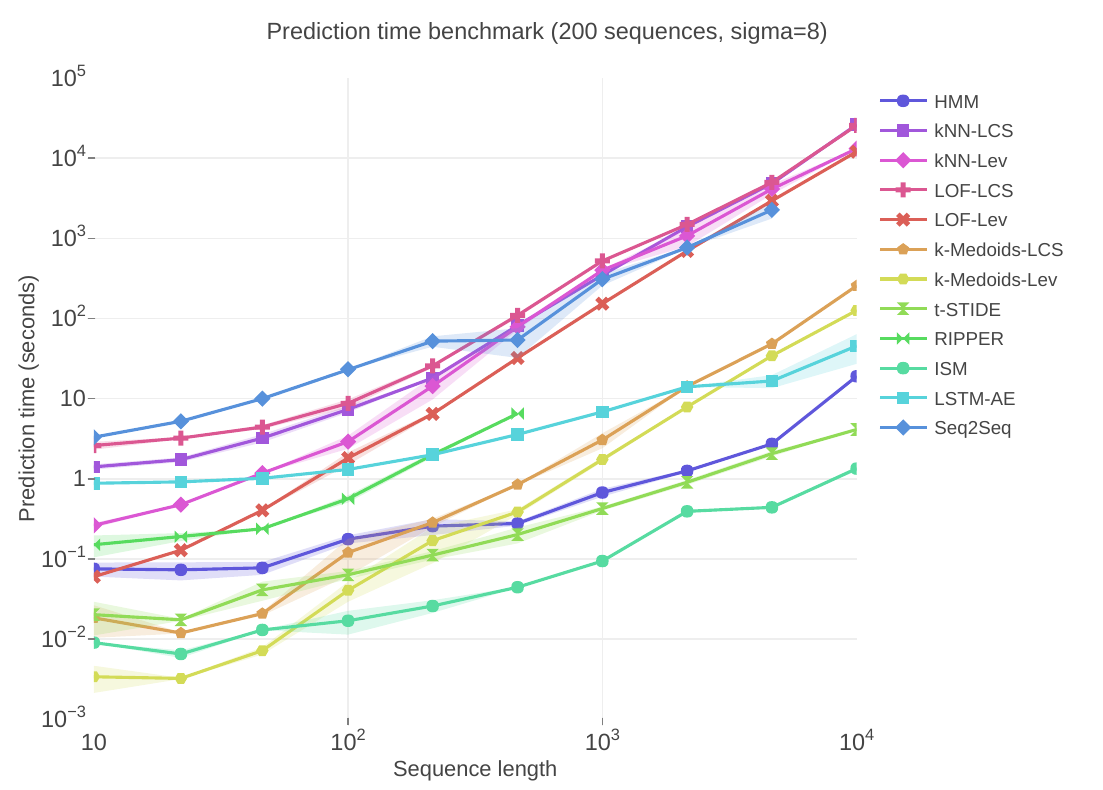}
\caption{Prediction time for increasing sequence length}
\label{fig:pred_seq_len}
\end{figure}

The computation time for training and prediction steps is reported in Figures \ref{fig:fit_samples} to \ref{fig:pred_seq_len}. While time measurements are impacted by hardware configuration (Sec. \ref{sec:exp:scala}), the slope of the curves and their ranking compared to other methods should remain the same for most running environments.

The measurements from Figures \ref{fig:fit_samples} and \ref{fig:pred_samples} show a poor scalability of algorithms relying on pairwise distance matrices, namely \lof, \knn and \kmedoids.
Most of the training and prediction time of these methods is dedicated to the computation of the distance matrix, and thus to the \lcs and Levenshtein algorithms.
Since training and testing sets have the same number of samples in this test, the previous assumption is confirmed by observing a similar training and prediction time for the methods.
In addition, \kmedoids is the only distance-based algorithm with a faster prediction time, caused by a smaller number of distances to compute.
The prediction step of this method requires only to compare a small number of medoids with the testing set, instead of performing a heavy pairwise comparison.
Regarding distance metrics, \lcs shows a much higher computation time than the Levenshtein distance despite a similar time complexity.
The resort to alternative and faster implementations\cite{crochemore2003speedinglcs,hunt1977fastlcs} is thus recommended.
Furthermore, parallel or distributed algorithms could be used for pairwise distance matrix computations, which would significantly reduce the computation time of these methods\cite{chang2008pairwise}.
However, distance-based methods would likely remain among the most computationally expensive algorithms when applied to a high number of observations.
Despite a very small $\sigma$, the rule-learning algorithm \ripper shows the highest training time, reaching our 12-hour timeout for 13,000 samples.
The missing results for the prediction step (Fig. \ref{fig:pred_samples}) are thus caused by an interrupted training.
On the opposite and as expected, the use of mini-batch learning by \lstmae and \seqtoseq allows the two methods to efficiently handle the increasing number of sequences, although we recommend to increase the batch size or the number of iterations according to the size of the training set.
However, such technique is only valid for the training step, and both methods show a scoring scalability comparable to the other algorithms.
The extreme simplicity of \tstide, which essentially stores subsequences in a dictionary at train time, makes this algorithm one of the fastest methods.
The increasing load does not affect much \ism, since the method stops iterating over the dataset if it does not find new interesting patterns after a given number of sequences.

We now use a fixed number of samples while increasing the length of the sequences and report the computation time in Figures \ref{fig:fit_seq_len} and \ref{fig:pred_seq_len}.
The careful reader will notice that both scalability tests, i.e. number of sequence-based and length-based, produce datasets containing the exact same number of symbols (e.g. $10^5 \ sequences * 20 \ symbols = 200 \ sequences * 10^4 \ symbols$).
This configuration reveals the true impact of samples and length on the scalability, while keeping the same volume of data.
While we still observe a poor scalability for distance-based algorithms caused by a high computation time to compute distances, the training and prediction time of these methods was reduced due to a smaller number of samples to handle by the core algorithm.
On the opposite, \ripper and \ism show a much higher training time when dealing with long sequences. However, the prediction time of these two methods only depends on the volume of data, i.e. the total number of symbols in the dataset, and will be impacted similarly by the number of samples and length.
Mini-batch methods are now subject to training batches of increasing volume, which reveals a poor scalability for \seqtoseq.
\lstmae performs better due to an early stopping mechanism, interrupting the training when the loss does not improve sufficiently over the iterations.
The computation time of these two neural networks could however be improved with the use of dedicated GPU architectures.
These tests show the limitations of \ripper, which suffers from a long training step, even for datasets of reasonable size.
Distance-based methods and \seqtoseq also show limited scalability, although \kmedoids provide fast predictions and \seqtoseq easily supports datasets containing a large number of samples.
\ism and \tstide show the best computation time for both training and prediction steps, and could even prove useful in lightweight applications.

\subsection{Memory usage}

\begin{figure}[!htb]
\centering
\includegraphics[width=0.9\linewidth]{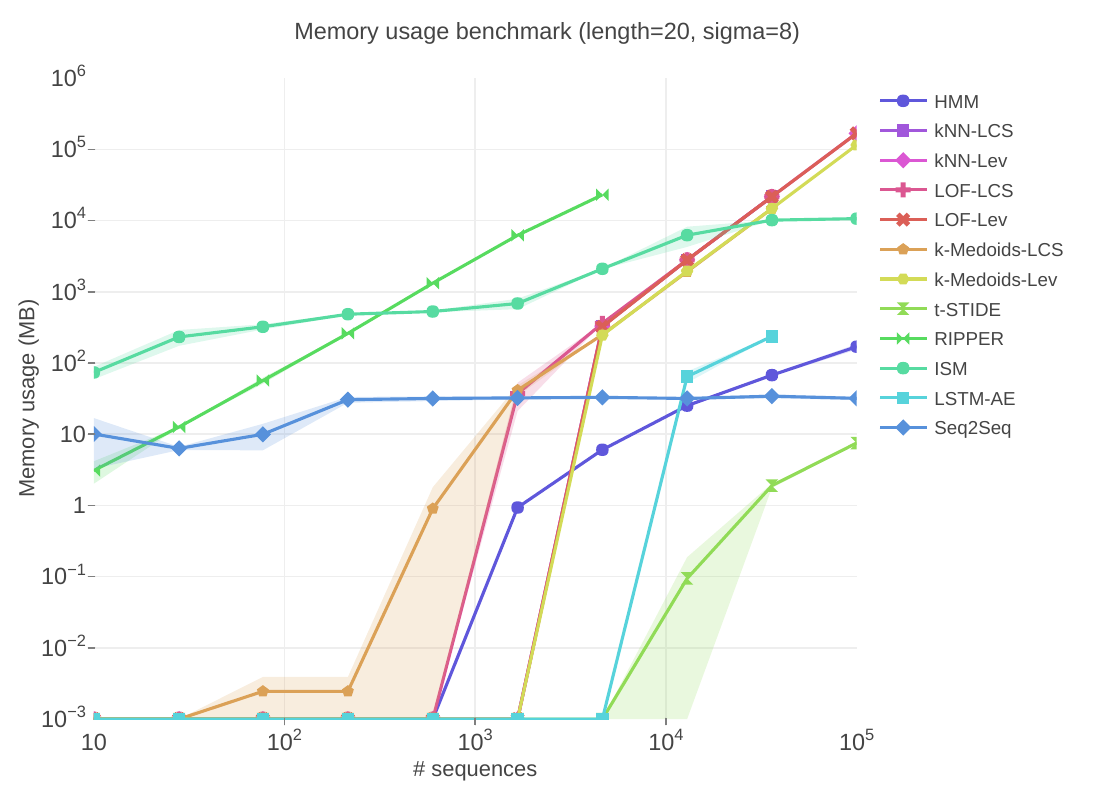}
\caption{Memory usage for increasing number of samples}
\label{fig:mem_samples}
\end{figure}

\begin{figure}[!htb]
\centering
\includegraphics[width=0.9\linewidth]{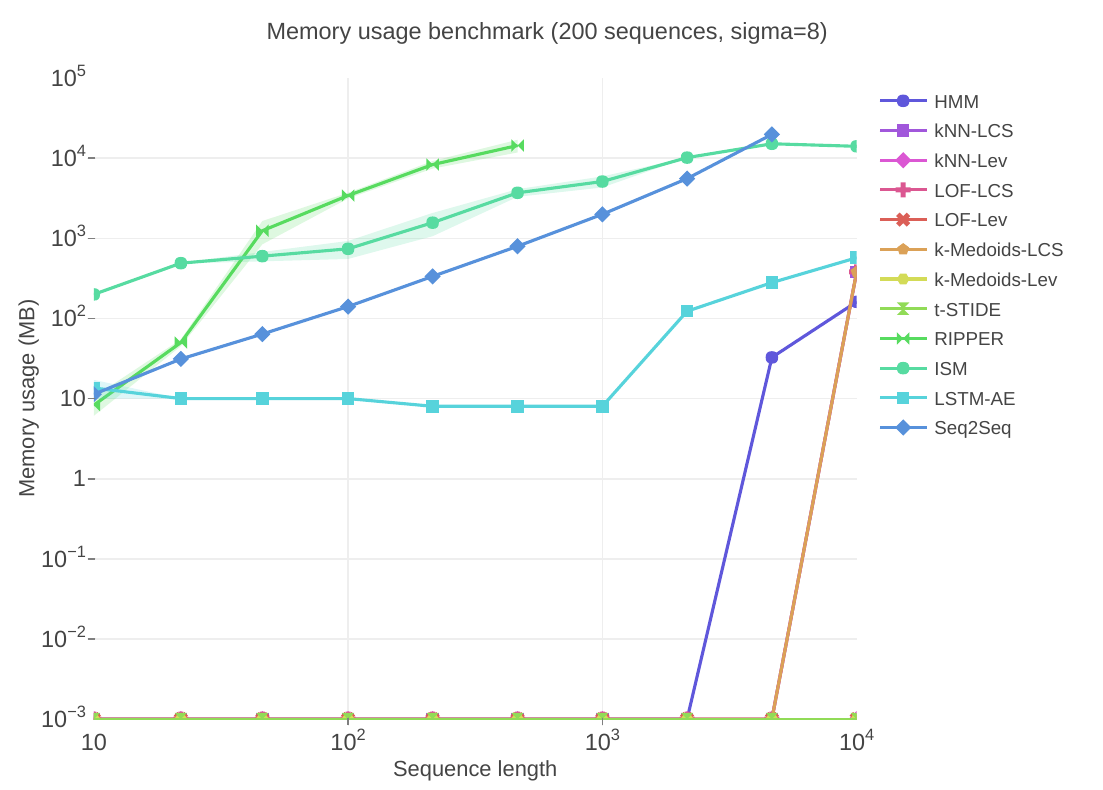}
\caption{Memory usage for increasing sequence length}
\label{fig:mem_seq_len}
\end{figure}

Monitoring the memory consumption in Figures \ref{fig:mem_samples} and \ref{fig:mem_seq_len} highlights important scalability constraints for several algorithms.

We first observe in Figure \ref{fig:mem_samples} that memory usage for \ripper and distance-based methods is strongly correlated with the number of input sequences.
\ripper shows a very high memory usage, although the method reaches our 12h timeout at train time before exceeding the limit of 256GB RAM.
Distance-based methods are also strongly impacted by the number of samples. However, most of the memory is here consumed by the pairwise distance matrix. Despite storage optimizations, e.g. symmetric matrix, integers are stored on 24 bytes by Python, resulting in a memory usage of 114GB and 167GB for \knnlev and \loflev, respectively.
Interestingly, \ism stabilizes at 10GB after having discovered a sufficient number of patterns from the data.
Mini-batch neural networks are not strongly impacted by the number of samples, and the small $\sigma$ limits the diversity of sequences, thus reducing the memory usage of \tstide.

The metrics reported in Figure \ref{fig:mem_seq_len} corroborate the previous conclusions.
The experiment reveals a number of rules learnt by \ripper increasing linearly with the number of events, the final model containing in average $\frac{\# events}{50}$ rules.
The size of the decision tree built by association rule learning is thus correlated with the volume of the data.
To the opposite, the memory usage of ISM stabilizes again after convergence, showing a more efficient internal representation of the data than \ripper.
The memory consumption of distance-based methods is very low due to small distance matrices, although the computation of \lcs shows a memory usage increasing with the length of the sequences compared.
Neural networks, especially \seqtoseq, are more impacted by the increasing sequence length. This is caused by a network topology depending on the size of the padded sequences, in addition to matrix multiplications of dimensionalities directly impacted by the length of the sequences.

We have observed that most algorithms have a memory consumption strongly related to the volume of input data.
The requirements of \ripper are too important for most systems, and distance-based methods are not suitable to address problems pertaining to more than 20,000 sequences.
Interestingly, we did not observe correlations between training or prediction time and memory usage, while one could expect fast algorithms consume more memory, performing faster computations due to a massive caching system.
If this may be true when comparing similar methods, the important differences in time and memory are here caused by major discrepancies in the approaches taken by the algorithms.

\subsection{Interpretability} \label{sec:interpretability}
The ability for humans to understand a machine learning model and the resulting predictions is called \textit{interpretability}.
This trait allows data scientists to validate the final model and provides useful insights on the targeted dataset, e.g. discovering valuable information about user behaviors which have an important business value.
While continuous scores are usually sufficient for automatic intervention modules, this information and the corresponding ranking may not be sufficient when a manual investigation of the anomalies is required.
This situation arises for critical applications, where false positives could strongly impact the brand image, e.g. deny access to services for a business partner, or incur heavy costs, e.g. component replacement based on failure prediction with applications to data centers and airplanes.
In this case, especially if many alerts are raised every day, the time allocated to manual investigation could be greatly reduced if we could provide the motivations behind high scores to the human expert.
Transparency is thus an essential criterion for the choice of algorithms in many applications, and data analysts may accept to trade performance for model accountability.
If human eyes may differentiate outlying activity from the underlying patterns in numerical time-series, this task is much harder for discrete event sequences, which emphasizes the need for model interpretability.

The internal representation of interpretable methods provides sufficient information to motivate a predicted score with respect to an input sequence.
For example, \hmm learns intuitive transition and emission matrices, providing an insightful weighted process flowchart.
Unusual event transitions in the test sequence can be visually highlighted by putting a threshold on the emission transition probabilities.
Pairwise distance matrices also convey valuable information and can be turned into intuitive visualizations. The matrices can be plotted as Voronoi diagrams, heat maps or fed into a multidimensional scaling (MDS) algorithm resulting in a scatter plot of chosen dimensionality.
If additional insight on the distance computations is required, \lcs is an intuitive metric and the subsequence common to two compared samples can be underlined. On the other hand, the cost matrix computed by Levenshtein is more difficult to read.
Further on, the scoring performed by distance-based methods can be easily motived in the previous 2D representations of distance matrices, e.g. by highlighting the test sample and its $k^{th}$ neighbor for \knn, or the corresponding medoid for \kmedoids.
The scoring function of \lof is more complex, as it studies the local density of a test sample and its neighbors.
Moving back to standard sequence representations, \tstide is extremely accountable and subsequences can be underlined based on their frequency in the model, thus motivating the resulting score.
Pointing out events incorrectly predicted by \ripper should also provide some information, and interesting patterns learnt by \ism could be emphasized similarly.
Neural networks are closer to black-box systems, and their interpretability has recently gained a lot of attention \cite{Zhang2018interpretability}.
However, recent efforts mostly focus on numerical and convolutional networks, which leaves room for future \lstm representations.
Differences between the input sequence and the reconstructed output could be highlighted for \seqtoseq, although it would not explain the underlying model.
For \lstmae, we could learn and plot a low dimensional numerical representation based on the internal representation of the network, but dimensionality reduction methods will often produce an output biased towards the average sample of the dataset \cite{onderwater2015outlier} and must be selected with care.
This is the reason why the reconstruction error is used with \seqtoseq to identify anomalies.

In order to overcome the lack of accountability of a given algorithm, an alternative approach is to infer meaningful rules based on the inputs and outputs predicted by a trained model\cite{fortuny2015extraction}.
The rule extraction method should provide simple rules showing a transparent decision, while minimizing the prediction error.
This is a popular approach used to improve the interpretability of classification models, in particular neural networks and support vector machines (\textsc{svm}s).
Two good rule extraction methods for classifiers are \textsc{osre}\cite{etchells2006extraction} and \textsc{hypinv}\cite{saad2009inversion}.
These methods are also compatible with novelty detection when the targeted model produces a binary output such as \textit{fraud} and \textit{non-fraud}.
If a continuous anomaly score is required to rank anomalies, we should then resort to regression rule extraction methods which learn rules producing a continuous output, e.g. REFANN\cite{setiono2002extraction}, ITER\cite{huysmans2006extraction} or classification and regression trees (\textsc{cart})\cite{breiman2017cart}.
Both regression and classification rule mining methods show good performance when applied to numerical or one-hot encoded input data.
In order to feed temporal data to these algorithms (or to any standard regression or classification methods), numerical features should be extracted from the sequences during a preprocessing step.
The feature selection must be performed with great care to minimize the amount of information lost, and was automated for continuous time-series in a previous work\cite{christ2016distributed}.
While different features should be selected for discrete event sequences, either manually or based on existing techniques\cite{wang2001feature,saidi2010feature}, any regression rule extraction technique can be subsequently applied for both data types.
The numerical latent representation provided by \lstm autoencoders could be used as input features for rule mining, but it would only improve the interpretability of the decoder, leaving aside the data transformation performed by the encoder.
Table \ref{tab:evaluation-summary} summarizes our observations about the scalability and interpretability of the methods surveyed.


\begin{table}
\centering
\caption{Scalability and interpretability summary. Runtime and memory consumption are reported for synthetic datasets of increasing number of samples and sequence length.}
\label{tab:evaluation-summary}
\resizebox{\columnwidth}{!}{
\begin{tabular}{llllll} \thickhline
& \multicolumn{2}{l}{\textbf{Training/prediction time}} & \multicolumn{2}{l}{\textbf{Mem. usage}} & \\
\textbf{Algorithm} & $\shortarrow{1}$ \textbf{Samples} & $\shortarrow{1}$ \textbf{Length} & $\shortarrow{1}$ \textbf{Samples} & $\shortarrow{1}$ \textbf{Length} & \textbf{Interpretability} \\ \thickhline
\hmm & Medium/Low & Low/Low & Low & Low & High \\
\knnlcs & High/High & Medium/High & High & Low & High \\
\knnlev & High/High & Medium/High & High & Low & Medium \\
\loflcs & High/High & Medium/High & High & Low & Medium \\
\loflev & High/High & Medium/High & High & Low & Medium \\
\kmedlcs & High/Low & Medium/Medium & High & Low & High \\
\kmedlev & High/Low & Medium/Medium & High & Low & Medium \\
\tstide & Low/Low & Low/Low & Low & Low & High \\
\ripper & High/Low & High/Medium & High & High & Medium \\
\ism & Low/Low & Medium/Low & Medium & Medium & High \\
\seqtoseq & Low/Medium & High/High & Low & High & Low \\
\lstmae & Low/Low & Low/Low & Low & Medium & Low \\ \thickhline
\end{tabular}
}
\end{table}

\section{Conclusions} \label{sec:conclusion}
This work studied the performance and scalability of state-of-the-art novelty detection methods based on a significant collection of real and synthetic datasets.
The standard metric used in the literature to compare event sequences is \lcs.
Given the evidence provided, we found that although \lcs produced more transparent insights than the Levenshtein distance, it did not exhibit better anomalies and was computationally more expensive.
Our experiments suggest that \knn, \kmedoids, \tstide and \lstmae are suitable choices to identify outliers in genomics, and that \hmm and \ripper are efficient algorithms to detect intrusions.
\hmm is a strong candidate for most novelty detection applications, and shows a good scalability and interpretability. These characteristics make \hmm appropriate for user behavior analysis, along with \knn, \kmedoids and \ism which also provide a good model accountability.
The fast scoring achieved by \hmm, \tstide and \ism implies an excellent management of heavy loads arising in production environments.
Major scalability constraints are pointed out for \ripper and distance-based methods, namely \knn, \kmedoids and \lof. The resort to alternative approaches when tackling large volumes of data is recommended.
The widely used \lstm networks show a lack of interpretability, and we believe that improving the understanding of recurrent networks as performed in \cite{Karpathy2016} would strongly benefit to the research community.
Most approaches evaluated in this study are suitable for supervised tasks based on event sequences. Studying how these methods compare in a supervised context would be of interest.

\section{Acknowledgments}
The authors wish to thank the Amadeus Middleware Fraud Detection team directed by Virginie Amar and J\'er\'emie Barlet, led by the product owner Christophe Allexandre and composed of Jean-Blas Imbert, Jiang Wu, Yang Pu and Damien Fontanes for building the \rights, \transfr and \transmo datasets.
MF gratefully acknowledges support from the AXA Research Fund.

\balance

\bibliographystyle{abbrv}
\bibliography{references}

\end{document}